\documentclass[12pt,draftcls,onecolumn]{IEEEtran}
\IEEEoverridecommandlockouts                              

\usepackage{amsmath,amssymb}

\DeclareMathOperator{\EX}{\mathbb{E}}

\usepackage{rotating}
\usepackage{chngcntr}
\usepackage{amsthm}
\usepackage{apptools}
\AtAppendix{\counterwithin{thm}{section}}
\usepackage{graphicx}
\usepackage{graphics}
\usepackage{amssymb}
\usepackage{amsmath}
\usepackage{color}
\usepackage{xspace}
\usepackage{bbm}
\usepackage{comment}
\usepackage{algorithm}
\usepackage{algpseudocode}
\usepackage{subfig}
\usepackage{psfrag}
\usepackage{cite}
\usepackage{tikz}
\usetikzlibrary{shapes,arrows}
\usetikzlibrary{positioning}

\tikzstyle{block}=[draw opacity=0.7,line width=1.4cm]

\definecolor{CranJ}{cmyk}{0,0.69,0.54,0.04} 
\definecolor{PinkJ}{cmyk}{0,0.71,0.43,0.12} 
\definecolor{Cran}{cmyk}{0,0.73,0.41,0.29} 
\definecolor{VRed}{cmyk}{0,0.75,0.25,0.2} 
\definecolor{ORed}{cmyk}{0,0.75,0.75,0} 
\definecolor{CBlue}{cmyk}{1,0.25,0,0} 

\title{\LARGE \bf  Projected Forward Gradient-Guided Frank-Wolfe Algorithm via Variance Reduction }

\author{Mohammadreza Rostami and Solmaz S. Kia, \emph{Senior Member, IEEE} %
  \thanks{The authors are with the Department of Mechanical and Aerospace Engineering, University of California Irvine, Irvine, CA 92697,  
    {\tt\small \{mrostam2,solmaz\}@uci.edu}.}%
}

\newcommand{\real}{{\mathbb{R}}}

\newcommand{\argmin}{\operatorname{argmin}}

\renewcommand{\baselinestretch}{.985}

\newcommand{\vect}[1]{\boldsymbol{\mathbf{#1}}}

 \newcommand{\boxend}{\hfill \ensuremath{\Box}}

\allowdisplaybreaks

\newtheorem{thm}{Theorem}[section]

\newtheorem{lem}{Lemma}[section]
\newtheorem{defn}{Definition}

\newtheorem{assump}{Assumption}

\newcommand{\oprocendsymbol}{\hbox{$\bullet$}}
\newcommand{\oprocend}{\relax\ifmmode\else\unskip\hfill\fi\oprocendsymbol}


\begin{document}
\fontsize{10}{11.1}\rm

\maketitle
\thispagestyle{empty}
\pagestyle{empty}

\begin{abstract}
This paper aims to enhance the use of the Frank-Wolfe (FW) algorithm for training deep neural networks. Similar to any gradient-based optimization algorithm, FW suffers from high computational and memory costs when computing gradients for DNNs. This paper introduces the application of the recently proposed projected forward gradient (Projected-FG) method to the FW framework, offering reduced computational cost similar to backpropagation and low memory utilization akin to forward propagation. Our results show that trivial application of the Projected-FG introduces non-vanishing convergence error due to the stochastic noise that the Projected-FG method introduces in the process. This noise results in an non-vanishing variance in the Projected-FG estimated gradient. To address this, we propose a variance reduction approach by aggregating historical Projected-FG directions. We demonstrate rigorously that this approach ensures convergence to the optimal solution for convex functions and to a stationary point for non-convex functions. Simulations demonstrate our results.

\end{abstract}

\section{Introduction}
Deep neural networks (DNNs) have become essential tools for solving function approximation problems over the past decade \cite{IG-YB-AC:16}. Training DNNs involves optimizing the parameter vector $\vect{\theta}\in\real^d$ (weights and biases) to minimize the loss between predictions and training data. To improve generalization, enhance interpretability, and reduce computational costs, it is desirable to promote sparsity in the DNN parameters. Traditionally, sparsity is achieved through post-training pruning. However, recent methods demonstrate that incorporating convex constraints directly into the training optimization problem enables one-shot pruning-aware training \cite{lu2022learning, zimmer2022compression}. These constraints impose sparsity and boundedness on $\vect{\theta}$~\cite[Table 1]{jaggi2013revisiting}. For constrained DNN training, the Frank-Wolfe (FW) algorithm \cite{pokutta2020deep} is often used, as it avoids costly projections onto constraint sets, making it suitable for high-dimensional~spaces. 

As with any first-order optimization method, the FW algorithm requires computing the gradient of the training cost function. However, computing the gradient in DNNs is not a trivial task; it demands special attention due to high-dimensional matrix multiplications and the efficient management of computational and memory resources needed for computing each layer's activations. This letter conducts a critical analysis of applying a newly proposed approximate gradient computation method with low computational and memory costs, known as the \emph{Projected Forward Gradient} (\emph{Projected-FG})\cite{silver2021learning, DS-AG-ID-MH-HH:22, baydin2022gradients, krylov2024moonwalk}, within the FW optimization framework.

\begin{figure}[t]
\centering
    \includegraphics[trim={10 15 10 22},clip,width=0.55\linewidth]{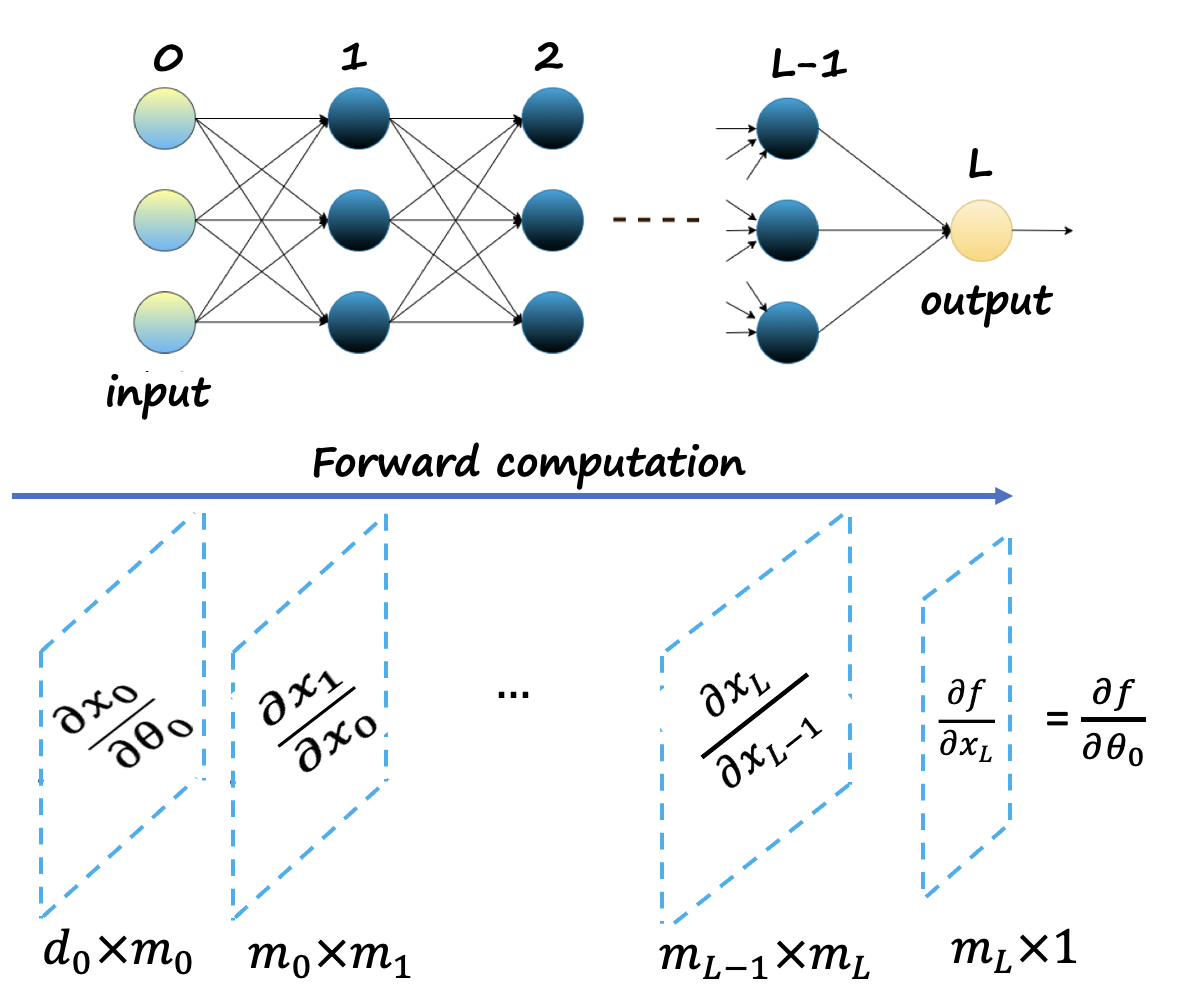}
\caption{{\small Using forward propagation to compute $\frac{\partial f}{\partial \theta_0}$ in a network with $L-1$ hidden layers, each with $n$ nodes.}}\vspace{-0.1in}
    \label{fig:forward_gradiant}
\end{figure}

For a DNN with cost $f(\vect{\theta})$ with $L+1$ layers and one output shown in Fig.~\ref{fig:forward_gradiant}, the gradient $\vect{g}(\vect{\theta})=\nabla f(\vect{\theta})=(\frac{\partial f}{\partial \vect{\theta}_0},\cdots, \frac{\partial f}{\partial \vect{\theta}_L})$, where $\vect{\theta}_i\in\real^{d_i}$ is the parameter vector of the $i$th layer, $i \in \mathcal{L}=\{0, 1, 2, ..., L\}$, is given by 
\begin{align}\label{eq::gradient_DNN}
   \vect{g}_i(\vect{\theta})= \underbrace{\frac{\partial \vect{x}_{i}}{\partial \vect{\theta}_i}}_{d_i\times m_i}\!\times\! \underbrace{\frac{\partial \vect{x}_{i+1}}{\partial \vect{x}_{i}}}_{m_i\times m_{i+1}} \!\times\!  \ldots \times \underbrace{\frac{\partial \vect{x}_L}{\partial \vect{x}_{L-1}}}_{m_{L-1}\!\times\! m_L}\!\times\! \underbrace{\frac{\partial f}{\partial \vect{x}_{L}}}_{m_L\times 1}.
\end{align}
Here, $\vect{g}_i(\vect{\theta})= \frac{\partial f}{\partial \vect{\theta}_{i}}$ and $\vect{x}_i=h_i(\vect{x}_{i-1},\vect{\theta}_i)\in\real^{m_i}$ is the activation (output) vector of $i$th layer and $\times$ denotes matrix multiplication. The gradient~\eqref{eq::gradient_DNN} in DNNs is typically computed using either \emph{forward propagation} (Fig. \ref{fig:forward_gradiant}) or \emph{backpropagation}. 
In forward propagation, starting from $\frac{\partial \vect{x}_{i}}{\partial \vect{\theta}_i}$, the gradient calculation involves multiplying \(L-1\) Jacobian matrices of dimensions \(m_i \times m_{i-1}\) from left to right up to the final term \(\frac{\partial f}{\partial \vect{x}_{L}}\), a vector of size \(m_{L-1}\). Although computationally expensive due to high-dimensional matrix multiplications, this method effectively utilizes memory since 
$\vect{x}_i$ of each layer $i \in \mathcal{L}$ can be computed sequentially and overwritten at the same memory location. Backpropagation starts at the output layer's $\frac{\partial f}{\partial \vect{x}_{L}}$ (a vector) and propagates backward, computing the gradient by backward multiplication between a vector and a high-dimensional matrix. While computationally efficient, backpropagation requires substantial memory to store all layer activations $\vect{x}_i$, $i \in \mathcal{L}$ \cite{novikov2022fewbitbackwardquantizedgradients}. 

Backpropagation is favored for its relatively low computational cost, and efforts have been made to reduce its memory usage. One approach involves checkpointing, where only a subset of activations is stored as checkpoints instead of saving every intermediate activation \cite{martens2012training, gruslys2016memory}. During the backward pass, these checkpoints are used to recompute the necessary intermediate activations that were not stored, reducing the memory footprint by performing recomputation on-the-fly. Another strategy is to eliminate the need to store activations altogether in reversible networks. In reversible networks, the input to each layer can be reconstructed from the output using the inverse function. Studies such as \cite{gomez2017reversible, mangalam2022reversible} exploit this property by recomputing activations backward during backpropagation. While these methods achieve memory savings, they come with an increased computational cost.

The \emph{Projected-FG}~\cite{silver2021learning, DS-AG-ID-MH-HH:22, baydin2022gradients, krylov2024moonwalk}  is a directional derivative computed in a forward fashion but with low computational cost:
\begin{defn}[Projected Forward Gradient]\label{def::fgradient}
    Give a function $f:\real^d\to\real$ the \emph{Projected-FG} is defined as defined as $\hat{\vect{g}}(\vect{\theta})=\langle \vect{g} (\vect{\theta}), \vect{u}\rangle\, \vect{u},$
 where $\vect{u} \in \mathbb{R}^d$ is a random vector with each entry $u_i$ independently and identically distributed (i.i.d) with zero mean and unit variance.\footnote{\noindent `Projected' here refers to the operation $\langle \vect{g} (\vect{\theta}), \vect{u}\rangle\vect{u}$, which projects the gradient $\vect{g} (\vect{\theta})$ on  vector $\vect{u}$. } \boxend
\end{defn}

 Using left-to-right forward multiplication starting with vector $\vect{u}_i \!\in \!\real^{d_i\times 1}$, $i \!\in\! \mathcal{L}$, results in subsequent multiplications between a row vector and a matrix in
 \begin{align*}
    \hat{\vect{g}}_i(\vect{\theta})= \! \Big(\underbrace{\vect{u}_i^\top}_{1\times d_i} \times \frac{\partial \vect{x}_{i}}{\partial \vect{\theta}_i}\!\times\! \frac{\partial \vect{x}_{i + 1}}{\partial \vect{x}_{i}} \times  \ldots \!\times\! \frac{\partial \vect{x}_L}{\partial \vect{x}_{L-1}}\times \frac{\partial f}{\partial \vect{x}_{L}}\Big)\!\times \!\vect{u}_i, 
\end{align*}
   providing the computational efficiency of backpropagation with the memory-efficient strategies of forward propagation.

\begin{figure}[t]
\centering
    \includegraphics[clip,width=0.55\linewidth]{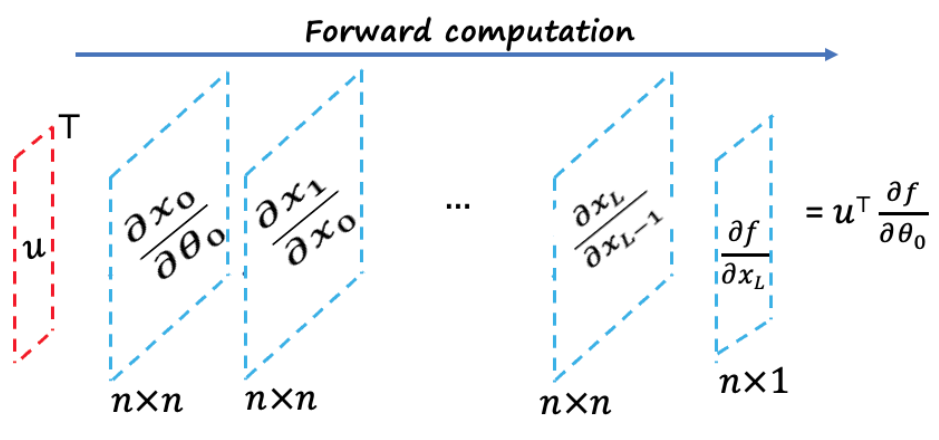}
    \caption{{\small Projected Forward Gradient }}
    \label{fig::fowrad and proj visul}
\end{figure}

However, the use of \textit{Projected-FG} in optimization algorithms is not straightforward, as  $\hat{\vect{g}}(\vect{\theta})$ is a noisy estimate of $\vect{g}(\vect{\theta})$. In this letter, we demonstrate that the straightforward application of the \textit{Projected-FG} method within the FW algorithm results in convergence to only a neighborhood of the optimal solution. We attribute this non-exact convergence to the non-vanishing variance in estimating the gradient with $\hat{\vect{g}}(\vect{\theta})$. To address the stochastic errors in gradient approximation, we propose an algorithm that reduces variance by averaging historical \textit{Projected-FG} directions. Inspired by momentum-based methods that accelerate gradient descent \cite{qian1999momentum}, our approach uniquely incorporates the \textit{Projected-FG} method to systematically reduce noise in gradient estimates. Unlike traditional momentum methods, which primarily use past gradients for faster convergence \cite{liu2018accelerating,liu2020accelerating, nakerst2020gradient}, our algorithm aggregates historical \textit{Projected-FG} directions, enhancing the convergence of the FW algorithm. Through rigorous proofs, we show that our method provides a biased variance-reduced \textit{Projected-FG} direction that converges to the true gradient with diminishing variance, ensuring convergence to the optimal solution for convex functions and to a stationary point for non-convex functions. A numerical example demonstrates the~results.

\section{Objective Statement}\label{setting}
Consider the constrained DNN training optimization problem 
\begin{equation}\label{problem} \min_{\vect{\theta}\in \mathcal{C}} f(\vect{\theta}),\end{equation} where  $\mathcal{C}\subset\real^d$ denotes a compact convex constraint set.  
\begin{assump}[Properties of the constraint set $\mathcal{C}$]\label{asm:comapct}
{ 
The convex set $\mathcal{C}$ is bounded with diameter $D\in\real_{>0}$, i.e., 
\begin{equation*}
    \|\vect{\theta} - \vect{\hat{\theta}}\| \leq D,\quad \forall\, \vect{\theta}, \vect{\hat{\theta}} \in \mathcal{C}.
\end{equation*}
} 
\end{assump}

\begin{assump}[Assumption on $M$-smoothness of the cost function]\label{assump::smoothness}
The cost function is $M$-smooth (or have an $M$-Lipschitz continuous gradient) in $\mathcal{C}$, i.e, there exists a constant $M \in\real_{>0}$ such that 
\begin{equation*}
    \|\vect{g}(\vect{\theta}) - \vect{g}(\vect{{\hat{\vect{\theta}}}})\| \leq M\|\vect{\theta} - \hat{\vect{\theta}}\|,\quad \forall \vect{\theta}, \hat{\vect{\theta}} \in\mathcal{C}.
\end{equation*}
\end{assump}
Recall that for a $M$-smooth $f(\vect{\theta})$, we have $f(\vect{\theta}) \leq f(\hat{\vect{\theta}}) + \langle \vect{g}(\hat{\vect{\theta}}), \vect{\theta} - \hat{\vect{\theta}} \rangle + \frac{M}{2} \|\vect{\theta} - \hat{\vect{\theta}}\|^2$, for any $\vect{\theta}, \hat{\vect{\theta}} \in \mathcal{C}
$\cite{nesterov2013introductory}.

To solve optimization problem~\eqref{problem}, Algorithm~\ref{FG-FW} presents the FW algorithm utilizing the \textit{Projected-FG} gradient estimation, where the computation of gradient 
$\vect{g}$ in line 5 is replaced by $\hat{\vect{g}}$. Our objective in this paper is to investigate the impact of using the \textit{Projected-FG} estimate of the gradient within the FW algorithm and devise strategies to counteract any performance degradation caused by the use of the noisy 
$\hat{\vect{g}}$.
To aid in our convergence analysis and algorithm design, we begin by reviewing some auxiliary results about the \textit{Projected-FG}.
\begin{lem}[Unbiasedness of the \it{PF-Gradient} \cite{baydin2022gradients}]\label{lem::unbias}
The \it{PF-Gradient} in Definition~\ref{def::fgradient}  is an unbiased estimate of $\vect{g} (\vect{\theta})$, i.e. $\EX[ \hat{\vect{g}}(\vect{\theta})] =\vect{g}(\vect{\theta})$.\boxend
\end{lem}
\begin{lem}[Upper bound on the \it{PF-Gradient}~\cite{baydin2022gradients}]\label{lemm::upperbound-FG} The \emph{PF-Gradient} in Definition~\ref{def::fgradient} satisfies
\begin{equation}\label{upper_bound_forward}
    \EX[ \,\|\hat{\vect{g}} (\vect{\theta})\|^2\,] \leq (n + 4)\| \vect{g} (\vect{\theta}) \|^2.
\end{equation}

\end{lem}
\begin{lem}[Variance of the \emph{PF-Gradient}]
\label{lem::lemm-variance-FG}
Under Assumptions~\ref{asm:comapct} and~\ref{assump::smoothness}, the \it{PF-Gradient} in Definition~\ref{def::fgradient} satisfies
\begin{equation}\label{variance}
    \EX[ \,\|\hat{\vect{g}} (\vect{\theta})-\vect{g}(\vect{\theta})\|^2\,] \leq \mathsf{G}=(n + 3)M^2D^2,\quad \forall\vect{\theta}\in\mathcal{C}. 
    \end{equation}
\end{lem}

In what follows, $f(\vect{\theta}^\star)$ is the optimal value of~\eqref{problem}. Moreover, the proofs of the formal results are presented in the appendix.

\begin{algorithm}[t]\small
\caption{Projected-FG Frank-Wolfe (FGFW)}\label{FG-FW}
\begin{algorithmic}[1]
\State \textbf{Require:} $\alpha_k$
\State \textbf{Input:} $\vect{\theta}_0 \in \mathcal{C}$
\State \textbf{Output:} $\vect{\theta}_{K+1}$
\For{$k = 1, 2, \dots, K$}
    \State Sample $\vect{u}_k \sim \mathcal{N}(\vect{0}, \mathit{I}_d)$ and compute $\hat{\vect{g}}(\vect{\theta}_k) = \langle \vect{g}(\vect{\theta}_k), \vect{u}_k \rangle \vect{u}_k$
    \State Compute $\vect{s}_k \in \argmin_{\vect{s} \in \mathcal{C}} \langle \hat{\vect{g}}(\vect{\theta}_k), \vect{s} \rangle$
    \State $\vect{\theta}_{k+1} = (1-\alpha_k)\vect{\theta}_k + \alpha_k\vect{s}_k$
\EndFor
\end{algorithmic}
\end{algorithm}

\section{Frank Wolfe Algorithm with Projected Forward Gradient}
Starting with $\vect{\theta}_0\in\mathcal{C}$, in each iteration, FW consults an oracle to identify a descent direction in $\mathcal{C}$ from $\vect{s}_k = \argmin_{\vect{s} \in\mathcal{C}}~\langle \vect{g}(\vect{\theta}_k), \vect{s}_k\rangle$. The process proceeds with the update $\vect{\theta}_{k+1} = (1 - \alpha_{k})\vect{\theta}_k + \alpha_k\vect{s}_k$, $\alpha_k\in[0,1]$, which is a convex combination of two vectors in $\mathcal{C}$ resulting in $\vect{\theta}_{k+1}$ remaining feasible within $\mathcal{C}$. Consequently, this method avoids projecting the updated point back into $\mathcal{C}$, thereby attributing a projection-free nature to the FW algorithm. The convergence of FW for convex functions is achieved by a rate of $O(1/k)$ and $O(1/\sqrt{k})$ for non-convex functions \cite{lacoste2016convergence, frank1956algorithm}. Next, we analyze the convergence behavior of Algorithm~\ref{FG-FW}, showing that incorporating \textit{Projected-FG} naively in FW framework introduces a convergence error. To simplify the study, we focus on smooth and convex costs.

\begin{lem}[Trajectories of Algorithm \ref{FG-FW} remain in $\mathcal{C}$]\label{projj_free}
Let $f$ in~\eqref{problem} be  differentiable. For any convex set $\mathcal{C}$, Algorithm \ref{FG-FW}, initialized at $\vect{\theta}_0 \in \mathcal{C}$ results in $\vect{\theta}_k \in \mathcal{C}$ for all $k\in\mathbb{Z}_{\geq0}$. 
\boxend
\end{lem}
Lemma~\ref{projj_free}'s proof follows the same reasoning as the original FW algorithm, using mathematical induction \cite{frank1956algorithm}. Since $\vect{\theta}_0\in\mathcal{C}$  and, by construction, 
$\vect{s}_k\in\mathcal{C}$ for all  $k\in\mathbb{Z}_{\geq0}$
  (see line 6 of Algorithm~\ref{FG-FW}), therefore, 
$\vect{\theta}_{k+1}$ , being a convex combination of two vectors in 
$\mathcal{C}$, will also reside in the convex set 
$\mathcal{C}$ for all 
$k\in\mathbb{Z}_{\geq0}$ . Lemma~\ref{projj_free} demonstrates that Algorithm~\ref{FG-FW} preserves the FW algorithm's ``interior point property." However, as we will show next, exact convergence is not maintained.

\begin{thm}[Convergence bound of Algorithm \ref{FG-FW} for convex functions]\label{thm::thm1}
Let $f$ in~\eqref{problem} be convex and Assumptions \ref{asm:comapct} and \ref{assump::smoothness} hold. Let $\alpha_k$ be a decaying function of $k$ with an order of $O(1/k)$ and $\psi_1=M D^2 \sqrt{n + 3}$. 
Then, Algorithm \ref{FG-FW} results in
\begin{align}\label{eq::PGFW_bound}
   \lim_{k\to\infty} \EX [f(\vect{\theta}_{k}) - f(\vect{\theta}^{\star})] = \frac{\psi_1}{1-\beta},
\end{align}
for $0\!<\!\beta\!<\!1$, with a convergence rate of order $O(1/k)$.\boxend
\end{thm}
Theorem~\ref{thm::thm1} shows Algorithm~\ref{FG-FW} converges with a sublinear rate to an error neighborhood of size $\psi_1$ of the optimal function value, indicating \textit{Projected-FG} cannot be trivially implemented within FW framework. To remove this convergence error, we propose Algorithm~\ref{alg::AFG-FW}, which we present next.

\section{Variance-Reduced Frank Wolfe Algorithm with Projected Forward Gradient}\label{methods} 
By examining the proof of Theorem~\ref{thm::thm1}, the error component $\psi_1$ is due to the second term in equation~\eqref{eq:::1}, i.e., $ \sqrt{\EX [\|\vect{g}(\vect{\theta}_k) - \hat{\vect{g}}(\vect{\theta}_k)\|^2]}$. This connects the convergence  error in Algorithm~\ref{FG-FW} to the non-vanishing variance of the \textit{Projected-FG} gradient estimator. To address this error, we introduced Algorithm~\ref{alg::AFG-FW}, which improved the convergence characteristics by estimating the gradient using aggregated historical \textit{F-Gradients}. This was achieved through
$\vect{v}_k \!=\! (1 - \gamma_k)\vect{v}_{k-1} + \gamma_k \hat{\vect{g}}(\vect{\theta}_k)$ in line 6 of Algorithm~\ref{alg::AFG-FW}. Rather than interpreting this line as a method for extrapolating the iterates \cite{liu2020improved, kidambi2018insufficiency}, we view $\vect{v}_k$ as a biased variance-reduced gradient estimator. 
This perspective is formally confirmed in Lemma~\ref{lem::variance_convergence} below. 

\begin{algorithm}[t]\small
\caption{Averaged Projected FG Frank Wolfe (AFGFW)}\label{alg::AFG-FW}
\begin{algorithmic}[1]
\State \textbf{Require:} $\alpha_k, \gamma_k$
\State \textbf{Input:} $\vect{\theta}_0 \in \mathcal{C}, \vect{v}_0 = \vect{0}$
\State \textbf{Output:} $\vect{\theta}_{K+1}$
\For{$k = 1, 2, \dots, K$}
    \State Sample $\vect{u}_k \sim \mathcal{N}(\vect{0}, \mathit{I}_d)$ and compute $\hat{\vect{g}}(\vect{\theta}_k) = \langle \vect{g}(\vect{\theta}_k), \vect{u}_k \rangle \vect{u}_k$
    \State Compute $\vect{v}_k = (1 - \gamma_k)\vect{v}_{k-1} + \gamma_k \hat{\vect{g}}(\vect{\theta}_k)$
    \State Compute $\vect{s}_k \in \argmin_{\vect{s} \in \mathcal{C}} \langle \vect{v}_k, \vect{s} \rangle$
    \State  $\vect{\theta}_{k+1} = (1-\alpha_{k})\vect{\theta}_k + \alpha_k\vect{s}_k$
\EndFor
\end{algorithmic}
\end{algorithm}

\begin{lem}[The variance of the averaged projected forward gradient estimator of Algorithm \ref{alg::AFG-FW} converges to zero]\label{lem::variance_convergence}
Consider optimization problem~\eqref{problem}. Let Assumption \ref{assump::smoothness} holds, and $\alpha_k$ to be a decaying function of $k$ with an order of $O(1/k)$, and $\eta_k = \frac{\alpha_k}{\gamma_k}$. Then, if $\lim_{k\to\infty}\!\eta_k\! = \!0$,
Algorithm~\ref{alg::AFG-FW} results in
\begin{align}
    \lim_{k\to \infty}\EX[\|\vect{v}_k - \vect{g}(\vect{\theta}_{k})\|^2] = 0,
\end{align}
with the rate of \mbox {$\min\{O(\eta_k^2), O(\gamma_k)\}$}. 
\boxend
\label{lem:averaged}
\end{lem}
The next results show that the asymptotic variance reduction of the gradient estimator $\vect{v}_k$
  leads to Algorithm \ref{alg::AFG-FW} achieving exact convergence. Here, we rely on the fact that since $\vect{v}_k\in\mathcal{C}$ a similar argument, as in Lemma \ref{projj_free}, can be established for Algorithm \ref{alg::AFG-FW}.

\begin{thm}[Convergence analysis of Algorithm \ref{alg::AFG-FW} for convex functions]\label{thm::thm2}
Let $f$ in~\eqref{problem} be convex and Assumptions \ref{asm:comapct} and \ref{assump::smoothness} hold. Let $\alpha_k$ be a decaying function of $k$ with an order of $O(1/k)$.
Then,  Algorithm~\ref{alg::AFG-FW} results in $ \lim_{k\to\infty} \EX [f(\vect{\theta}_{k}) - f(\vect{\theta}^{\star})] \to 0$ with a sublinear convergence rate of  order
{$\min \{O(\frac{\alpha_k}{\gamma_k}), O(\gamma_k)\}$}.
\boxend
\end{thm}

According to Theorem \ref{thm::thm2}, by selecting $\gamma_k = O(1/\sqrt{k})$,  Algorithm \ref{alg::AFG-FW} achieves its optimal convergence rate of $O(1/\sqrt{k})$. This improves over Algorithm \ref{FG-FW} by reducing noise through the incorporation of past \textit{Projected-FG} directions.

In DNNs, often the cost function is non-convex. The following theorem presents the convergence guarantee of Algorithm~\ref{alg::AFG-FW} for non-convex functions. For non-convex constrained optimization problems, the convergence criterion typically used is the so-called \emph{FW gap}, defined as 
$$
\mathcal{K}(\vect{\theta}) = \max_{\vect{s} \in  \mathcal{C}} \langle \vect{s} - \vect{\theta}, -\vect{g}(\vect{\theta}) \rangle,
$$
where $\mathcal{K}(\vect{\theta}) = 0$ if and only if $\vect{\theta}$ is a stationary point \cite{lacoste2016convergence}. 

\begin{thm}[Convergence analysis of Algorithm \ref{alg::AFG-FW} for non-convex functions]\label{thm::thm2_notcon}
Consider Algorithm \ref{alg::AFG-FW} under Assumptions \ref{asm:comapct} and \ref{assump::smoothness}. Let $\alpha_k$ and $\gamma_k$ be decaying functions of $k$ with orders of $O(1/k)$ and $O(1/\sqrt{k})$, respectively. Then, $ \lim_{K \to \infty} \EX [\mathcal{K}(\vect{\theta}_K)] = 0$ with an order of $O\left(\frac{1}{\log K}\right)$, satisfying the upper bound 
\begin{align*}
 \EX \left[\mathcal{K}\left(\vect{\theta}_K\right)\right]
& \leq \frac{f\left(\vect{\theta}_0\right)-f\left(\vect{\theta}^\star\right)}{\log K}  + E,\quad k \in \{1, 2, \dots, K\},
\end{align*}
where $f(\vect{\theta}^\star)$ is the optimal value of problem \eqref{problem} and $E$ is a decaying term of the order $O\left(\frac{1}{\log K}\right)$.
\boxend
\end{thm}

Theorem \ref{thm::thm2_notcon} shows Algorithm \ref{alg::AFG-FW} achieves a convergence rate of $O(\frac{1}{\log K})$ to a stationary point for non-convex functions.

\section{Numerical Evaluation}\label{numerical}
We evaluate the training loss and memory reduction of Algorithms \ref{FG-FW} and \ref{alg::AFG-FW} when training a DNN with constrained parameters for a multinomial classification task. The data set we use, the MNIST dataset~\cite{deng2012mnist}, consists of $28\times28$ pixel grayscale images of handwritten digits ranging from $0$ to $9$. Here, the purpose of training is to classify each digit correctly, aiming to create a model that can precisely predict the digit represented in any input image. We utilize a DNN with a depth of $5$ layers, each containing $10$ neurons. A ReLU activation function is applied across all layers. The model is trained for $50$ epochs. To encourages sparsity in the model, we use an $\ell_1$-norm constraint on the parameters in all layers, whose bound is set at $10^{-4}$. 
 To implement Algorithms \ref{FG-FW} and \ref{alg::AFG-FW}, we choose $\gamma_k = \frac{1}{\sqrt{k}}$ and $\alpha_k = \frac{1}{k}$. For Algorithm 1, we set $\alpha_k = \frac{2}{k + 2}$ which is a standard step size in the FW Algorithm.

\begin{figure}[t]
\centering
    \includegraphics[scale=0.29]{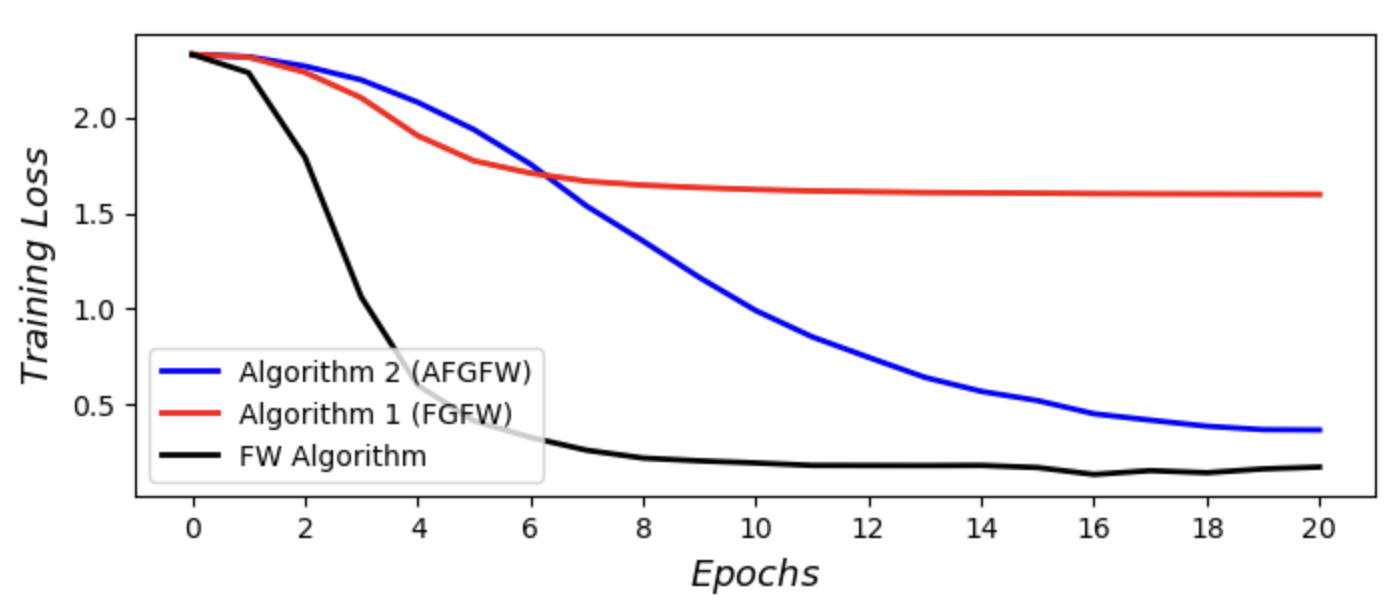}\\{ \small (a): Training loss }\\
     \includegraphics[scale=0.29]{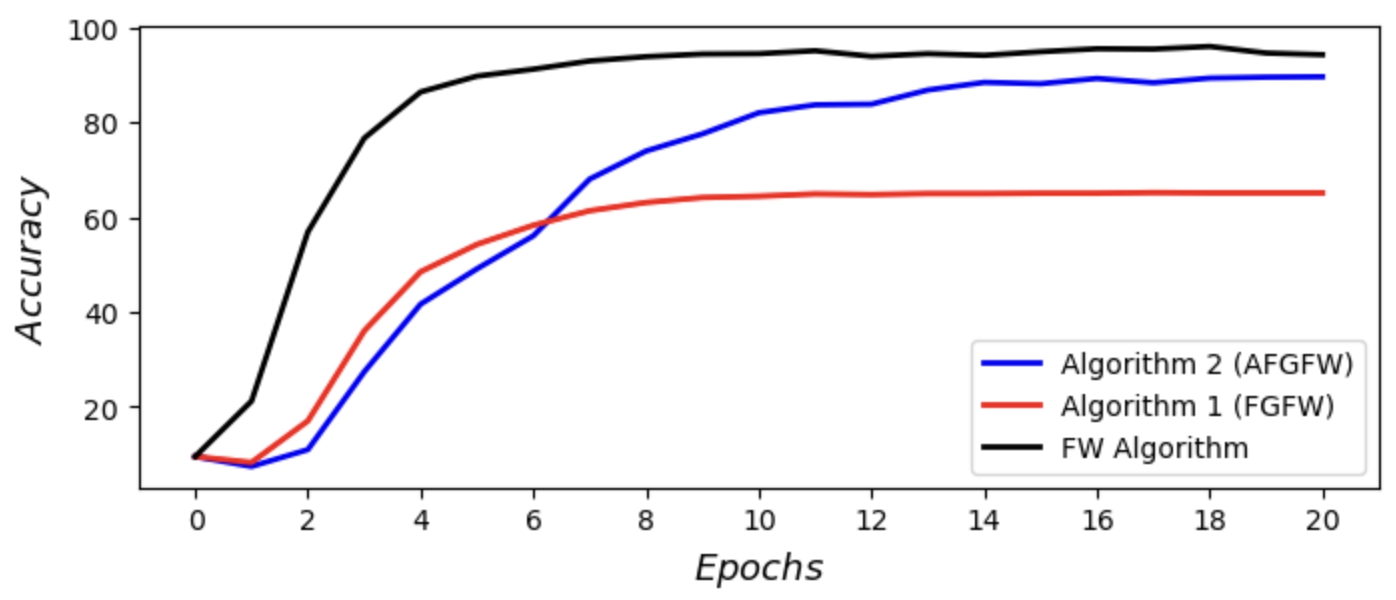}\\\small (b): Accuracy \vspace{-0.05in}
    \caption{{\small Training loss and accuracy with respect to the number of epochs for Algorithm \ref{FG-FW} and \ref{alg::AFG-FW} over
20 epochs}.}
    \label{loss_plot}
\end{figure}

Figure \ref{loss_plot}(a) shows the training loss for the FW algorithm and Algorithms \ref{FG-FW} and \ref{alg::AFG-FW}. As predicted by Theorem~\ref{thm::thm1}, the naive adaptation of the \textit{Projected-FG} method within the FW framework in Algorithm \ref{FG-FW} results in a steady-state error due to the non-vanishing noise in the gradient estimate  \(\hat{\vect{g}}(\vect{\theta}_k)\).

On the other hand, Algorithm \ref{alg::AFG-FW} shows convergence behavior as proved in Theorem \ref{thm::thm2_notcon}, albeit slower than Algorithm FW algorithm. But, FW algorithm achieves faster convergence by using the full gradient computed via backpropagation, which requires higher memory usage. The maximum memory usage during training with \textit{Projected-FG} in both Algorithms \ref{FG-FW} and \ref{alg::AFG-FW} is approximately 990 MiB $\pm$ 10\% (where 1 mebibyte (MiB) = 1024 kibibytes (KiB)) per epoch. In contrast, using backpropagation to calculate the gradient results in a memory usage of about 2150 MiB $\pm$ 10\% per epoch. This demonstrates the advantage of \textit{Projected-FG} methods in training DNNs by significantly reducing memory consumption compared to the backpropagation method used in FW algorithm. 

Figure \ref{loss_plot}(b) shows the test dataset accuracy for the FW algorithm, Algorithm \ref{FG-FW}, and Algorithm \ref{alg::AFG-FW}, with accuracies of $94.29\%$, $65.13\%$, and $89.63\%$, respectively, after $20$ epochs. Notably, enforcing the $\ell_1$-norm constraint sparsified the DNN parameters, setting $2,387$ of $8,290$ elements of the optimized $\vect{\theta}$ to zero, which improved generalization to unseen datasets. Without this constraint, accuracy drops to $86.31\%$, highlighting the benefits of constraints in DNN training.

\section{Conclusion}\label{sec::conclu}
This paper presented an enhanced approach to the FW algorithm for training DNNs by integrating the \emph{Projected-FG} method. Our study demonstrated that while the straightforward application of \emph{Projected-FG} within the FW framework reduced computational and memory costs, it introduced non-vanishing convergence errors due to stochastic noise. To address this challenge, we proposed a novel algorithm that incorporated variance reduction by averaging historical \emph{Projected-FG} directions. This approach not only mitigated the noise but also ensured convergence to the optimal solution for convex functions and to a stationary point for non-convex functions. Through theoretical analysis and numerical simulations, we validated the effectiveness and efficiency of our proposed method. The results highlighted the potential of using \emph{Projected-FG} to achieve memory-efficient training of DNNs without compromising convergence properties. Future work will explore the proposed method in distributed settings aligned with network topologies. We will  also investigate  using other distributions for sampling $\vect{u}_k$, including the symmetric and binary Rademacher distribution, known for its variance reduction attribute. The bounded nature of the Rademacher variables means that they are less susceptible to extreme outliers compared to the normal distribution. This can lead to more stable performance in algorithms as demonstrated for example in \cite{spall1992multivariate, rostami2024fedscalar}.

\bibliographystyle{ieeetr}%
\bibliography{bib/alias,bib/Reference}
\section*{Appedix}
\begin{proof}[Proof of Lemma~\ref{lem::lemm-variance-FG}] We note that
    $$ \EX[ \,\|\hat{\vect{g}} (\vect{\theta})-\vect{g}(\vect{\theta})\|^2\,]\stackrel{\text{(a)}}{=} \EX[ \,\|\hat{\vect{g}} (\vect{\theta})\|^2]-\|\vect{g}(\vect{\theta})\|^2\stackrel{\text{(b)}}{\leq } $$
    $$ (n + 3)\| \vect{g} (\vect{\theta}) \|^2 \stackrel{\text{(c)}}{\leq}  (n + 3)M^2 \|\vect{\theta}-\vect{\theta}^\star\|^2\stackrel{\text{(d)}}{\leq} (n + 3)M^2D^2.$$
Right hand side (RHS) of equality (a) follows from Lemma~\ref{lem::unbias}, RHS of inequality (b) follows from Lemma~\ref{lemm::upperbound-FG}, RHS of inequality (c) follows from Assumption~\ref{assump::smoothness} and RHS of inequality (d) follows from Assumption~\ref{asm:comapct}.
\end{proof}

\begin{proof}[Proof of Theorem~\ref{thm::thm1}]
Given Lemma~\ref{projj_free} that ensures $\vect{\theta}_k\in\mathcal{C}$ and using Assumption \ref{assump::smoothness}, we can write
$\EX[f(\vect{\theta}_{k+1})|\vect{\theta}_{k}] \leq f(\vect{\theta}_{k})+\big \langle\vect{g}(\vect{\theta}_{k}), \EX[\vect{\theta}_{k+1} - \vect{\theta}_{k}|\vect{\theta}_{k}] \big \rangle +\frac{M}{2}\EX\big[\|\vect{\theta}_{k+1} - \vect{\theta}_{k}\|^2\big| \vect{\theta}_{k}\big].$
Then, $\vect{\theta}_{k+1} = (1-\alpha_{k})\vect{\theta}_k + \alpha_k\vect{s}_k$ 
results in 
\begin{align}\label{eq::3}
     &\EX[f(\vect{\theta}_{k+1})|\vect{\theta}_{k}]  \leq f(\vect{\theta}_{k})+\alpha_k \big \langle \vect{g}(\vect{\theta}_k), \EX[\vect{s}_{k} - \vect{\theta}_{k}|\vect{\theta}_{k}] \big \rangle  
    +\frac{M\alpha_k ^2}{2}\EX\big[\|\vect{s}_{k} - \vect{\theta}_{k}\|^2 \big| \vect{\theta}_{k}\big]\nonumber \\
    &\stackrel{\text{(a)}}{=} f(\vect{\theta}_{k})+\alpha_k \big \langle \vect{g}(\vect{\theta}_k) - \hat{\vect{g}}(\vect{\theta}_k), \EX[\vect{s}_{k} - \vect{\theta}_{k}|\vect{\theta}_{k}] \big \rangle  + \alpha_k \big \langle \hat{\vect{g}}(\vect{\theta}_k), \EX[\vect{s}_{k} - \vect{\theta}_{k}|\vect{\theta}_{k}] \big \rangle 
    +\frac{M \alpha_k ^2}{2}\EX\big[\|\vect{s}_{k} - \vect{\theta}_{k}\|^2 \big| \vect{\theta}_{k}\big],
\end{align}
where RHS of equality (a) in~\eqref{eq::3} follows from adding and subtracting $\alpha_k \langle \hat{\vect{g}}(\vect{\theta}_k), \EX[\vect{s}_{k} - \vect{\theta}_{k}|\vect{\theta}_{k}] \big \rangle$.
Applying total expectation on both sides of \eqref{eq::3} we have
\begin{align}\label{eq::4}
    \EX &[f(\vect{\theta}_{k+1})]  \leq \EX [f(\vect{\theta}_{k})] + \alpha_k \EX [\langle \vect{g}(\vect{\theta}_k) - \hat{\vect{g}}(\vect{\theta}_k), \vect{s}_{k} - \vect{\theta}_{k} \rangle]  + \alpha_k \EX [\langle \hat{\vect{g}}(\vect{\theta}_k), \vect{s}_{k} - \vect{\theta}_{k} \rangle] + \frac{M \alpha_k^2}{2} \EX \big[\|\vect{s}_{k} - \vect{\theta}_{k}\|^2 \big] \nonumber \\
    &\stackrel{\text{(a)}}{\leq} \EX [f(\vect{\theta}_{k})] + \alpha_k \EX [\langle \vect{g}(\vect{\theta}_k) - \hat{\vect{g}}(\vect{\theta}_k), \vect{s}_{k} - \vect{\theta}_{k} \rangle] + \alpha_k \EX [\langle \hat{\vect{g}}(\vect{\theta}_k), \vect{\theta}^{\star} - \vect{\theta}_{k} \rangle] + \frac{M \alpha_k^2}{2} \EX \big[\|\vect{s}_{k} - \vect{\theta}_{k}\|^2 \big],
\end{align}

where in RHS of inequality (a) in~\eqref{eq::4}, $\left\langle \vect{s}_k, \hat{\vect{g}}(\vect{\theta}_k)\right\rangle$ is replaced by its upper bound $\left\langle\vect{\theta}^{\star}, \hat{\vect{g}}(\vect{\theta}_k)\right\rangle$ because $\left\langle\vect{\theta}^{\star}, \hat{\vect{g}}(\vect{\theta}_k)\right\rangle \geq \min _{\vect{s} \in \mathcal{C}}\left\{\left\langle \vect{s}, \hat{\vect{g}}(\vect{\theta}_k)\right\rangle\right\}=\left\langle \vect{s}_k, \hat{\vect{g}}(\vect{\theta}_k)\right\rangle$.
Adding and subtracting $\alpha_k \EX [\langle \vect{g}(\vect{\theta}_k), \vect{\theta}^{\star} - \vect{\theta}_k\rangle]$ in \eqref{eq::4} results in 
\begin{align}
    \EX &[f(\vect{\theta}_{k+1})] \leq \EX [f(\vect{\theta}_{k})] + \alpha_k \EX [\langle \vect{g}(\vect{\theta}_k) - \hat{\vect{g}}(\vect{\theta}_k), \vect{s}_{k} - \vect{\theta}^{\star} \rangle]  + \alpha_k \EX [\langle \vect{g}(\vect{\theta}_k), \vect{\theta}^{\star} - \vect{\theta}_{k} \rangle] + \frac{M \alpha_k^2}{2} \EX \big[\|\vect{s}_{k} - \vect{\theta}_{k}\|^2 \big], \nonumber \\
    &\stackrel{\text{(a)}}{\leq} \EX [f(\vect{\theta}_{k})] + \alpha_k \EX [\langle \vect{g}(\vect{\theta}_k) - \hat{\vect{g}}(\vect{\theta}_k), \vect{s}_{k} - \vect{\theta}^{\star} \rangle]  + \alpha_k \EX [f(\vect{\theta}^{\star}) - f(\vect{\theta}_k)] + \frac{M \alpha_k^2}{2} \EX \big[\|\vect{s}_{k} - \vect{\theta}_{k}\|^2 \big], \nonumber \\
    &\stackrel{\text{(b)}}{\leq} \EX [f(\vect{\theta}_{k})] + \alpha_k D \EX [\|\vect{g}(\vect{\theta}_k) - \hat{\vect{g}}(\vect{\theta}_k)\|]  + \alpha_k \EX [f(\vect{\theta}^{\star}) - f(\vect{\theta}_k)] + \frac{M \alpha_k^2 D^2}{2}. \label{eq::42}
\end{align}

RHS of inequality (a) in \eqref{eq::42} follows from the convexity property of the cost function $f$~\big(i.e., $f(\vect{\theta}^{\star}) - f(\vect{\theta}_k) \geq \langle \vect{g}(\vect{\theta}_k), ~\vect{\theta}^{\star} - \vect{\theta}_k\rangle$\big), RHS of inequality (b) in \eqref{eq::42} follows from Cauchy-Schwarz inequality and Assumption \ref{asm:comapct}. Adding and subtracting $f(\vect{\theta}^{\star})$ on the both sides of \eqref{eq::42}, we get
\begin{align*}
    \EX [f(\vect{\theta}_{k+1}) - f(\vect{\theta}^{\star})] 
    &\leq (1 - \alpha_k) \EX [f(\vect{\theta}_k) - f(\vect{\theta}^{\star})]  + \alpha_k D \EX [\|\vect{g}(\vect{\theta}_k) - \hat{\vect{g}}(\vect{\theta}_k)\|] + \frac{M \alpha_k^2 D^2}{2}.
\end{align*}

It follows from Lyapunov's inequality \cite{YN:18} that  $\EX [\|\vect{g}(\vect{\theta}_k) - \hat{\vect{g}}(\vect{\theta}_k)\|]\leq \sqrt{\EX [\|\vect{g}(\vect{\theta}_{k}) - \hat{\vect{g}}(\vect{\theta}_k)\|^2]}$; therefore, using the bound in Lemma~\eqref{lem::lemm-variance-FG}, we can write 
\begin{align}\label{eq::1}
    \EX [f(\vect{\theta}_{k+1}) - f(\vect{\theta}^{\star})] 
    &\leq (1 - \alpha_k) \EX [f(\vect{\theta}_k) - f(\vect{\theta}^{\star})]  + \alpha_k M D^2 \sqrt{n+3} + \frac{M \alpha_k^2 D^2}{2}.
\end{align}

By defining $\delta_{k+1} = \EX [f(\vect{\theta}_{k+1}) - f(\vect{\theta}^{\star})]$, then, we have
\begin{align}\label{recur}
    \delta_{k+1} &\leq (1-\alpha_k) \delta_{k} + \psi_1 \alpha_k  + \psi_2 \alpha_k^2,\quad \forall k \geq 0,\nonumber 
\end{align}
where $\psi_1 \!= \!M D^2 \sqrt{n + 3}$ and $\psi_2\! =\! \frac{M D^2}{2}$. Invoking Comparison Lemma~\cite{HKK:02} and Lyapunov convergence analysis, we consider $V_k\! =\! |\delta_k|$ as our candidate Lyapunov function to~write
\begin{align*}
    V_{k+1} - V_k &= |(1-\alpha_k) \delta_{k} + \alpha_k(\psi_1 + \psi_2\alpha_k)| - |\delta_k| \nonumber \\
    & \leq -\alpha_k |\delta_{k}| + \alpha_k(\psi_1 + \psi_2\alpha_k)  \nonumber \\
    & = -\beta \alpha_k |\delta_{k}| - (1-\beta) \alpha_k |\delta_{k}| + \alpha_k(\psi_1 + \psi_2\alpha_k),
\end{align*}
where $0< \beta < 1$. Then if $|\delta_{k}| \geq \frac{\psi_1 + \psi_2\alpha_k}{1-\beta}$, we have exponentially convergence to the ball with radius $\frac{\psi_1 + \psi_2\alpha_k}{1-\beta}$. Therefore, we can write $
   \lim_{k\to\infty} \EX [f(\vect{\theta}_{k}) - f(\vect{\theta}^{\star})] = \frac{\psi_1}{1-\beta},$ with the rate determined by $O(\alpha_k)$.
\end{proof}

\begin{proof}[Proof of Lemma~\ref{lem::variance_convergence}]  In Algorithm~\ref{alg::AFG-FW}, we have 
\begin{align}
   \|&\vect{v}_k - \vect{g}(\vect{\theta}_k)\|^2 = \|(1 - \gamma_k)\vect{v}_{k-1} + \gamma_k \hat{\vect{g}}(\vect{\theta}_k) - \vect{g}(\vect{\theta}_k)\|^2 \nonumber \\
   & = \|(1 - \gamma_k)(\vect{v}_{k-1} -\vect{g}(\vect{\theta}_k)) + \gamma_k(\hat{\vect{g}}(\vect{\theta}_k) - \vect{g}(\vect{\theta}_k))\|^2 \nonumber \\
   &= (1 - \gamma_k)^2 \|(\vect{v}_{k-1} -\vect{g}(\vect{\theta}_k))\|^2 + \gamma_k^2 \| (\hat{\vect{g}}(\vect{\theta}_k) - \vect{g}(\vect{\theta}_k))\|^2 + 2 \gamma_k(1 - \gamma_k)(\vect{v}_{k-1} -\vect{g}(\vect{\theta}_k))^\top(\hat{\vect{g}}(\vect{\theta}_k) - \vect{g}(\vect{\theta}_k))\nonumber \\
   &= (1 - \gamma_k)^2 \|\vect{v}_{k-1} -\vect{g}(\vect{\theta}_k) ~+ \vect{g}(\vect{\theta}_{k-1}) - \vect{g}(\vect{\theta}_{k-1}) \|^2 + \gamma_k^2 \| (\hat{\vect{g}}(\vect{\theta}_k) - \vect{g}(\vect{\theta}_k))\|^2 \nonumber \\
   &~~+ 2 \gamma_k(1 - \gamma_k)(\vect{v}_{k-1} -\vect{g}(\vect{\theta}_k))^\top (\hat{\vect{g}}(\vect{\theta}_k) - \vect{g}(\vect{\theta}_k)).
\end{align}

Conditional expectation on both sides leads to
\begin{align}\label{final::expec}
    \EX[\|\vect{v}_k - \vect{g}(\vect{\theta}_k)&\|^2|\boldsymbol{\sigma}_{0:k}] = (1 - \gamma_k)^2 \EX[\|\vect{v}_{k-1} -\vect{g}(\vect{\theta}_k) + \vect{g}(\vect{\theta}_{k-1}) - \vect{g}(\vect{\theta}_{k-1}) \|^2|\boldsymbol{\sigma}_{0:k}]  \nonumber \\
   &+ \gamma_k^2\EX[ \| (\hat{\vect{g}}(\vect{\theta}_k) - \vect{g}(\vect{\theta}_k))\|^2|\boldsymbol{\sigma}_{0:k}],
\end{align}
where $\boldsymbol{\sigma}_{0:k}$ is a vector containing all the randomness up to iteration $k$. 

Express \(\vect{g}(\vect{\theta}_{k-1}) - \vect{g}(\vect{\theta}_k) = \zeta_k \vect{p}_k\), where \(\vect{p}_k\) is a normalized vector (\(\|\vect{p}_k\| = 1\)) and \(\zeta_k \in \mathbb{R}_{>0}\). Next note that, from line 8 of Algorithm~\ref{alg::AFG-FW} we have \(\|\vect{\theta}_k - \vect{\theta}_{k-1}\| = \alpha_k \|\vect{s}_k - \vect{\theta}_k\| \leq \alpha_k D\) (recall line 8). Then, under Assumption~\ref{assump::smoothness} we can write $ \|\vect{g}(\vect{\theta}_k) - \vect{g}(\vect{\theta}_{k-1})\| \leq M \|\vect{\theta}_k - \vect{\theta}_{k-1}\| \leq M \alpha_k D.$ Thus, \(\lim_{k \to \infty} \zeta_k = 0\) at least as fast as \(\alpha_k\).

Hence, if we choose $\alpha_k$ and $\gamma_k$ such that $\lim_{k\to\infty}\frac{\alpha_k}{\gamma_k} = 0$, then, we can write $\vect{g}(\vect{\theta}_{k-1}) - \vect{g}(\vect{\theta}_k) = \eta_k\gamma_k \vect{p}_k$ where $\eta_k = \frac{\alpha_k}{\gamma_k}\in\real_{>0}$ and $\lim_{k\to\infty}\eta_k=0$. Lastly, taking total expectation on both sides of \eqref{final::expec}, we have
\begin{align}\label{eq::13}
    \EX&[\|\vect{v}_k - \vect{g}(\vect{\theta}_k)\|^2] = (1 - \gamma_k)^2 \EX[\|\vect{v}_{k-1} - \vect{g}(\vect{\theta}_{k-1}) + \eta_k\gamma_k \vect{p}_k\|^2] + \gamma_k^2\EX[ \| (\hat{\vect{g}}(\vect{\theta}_k) - \vect{g}(\vect{\theta}_k))\|^2] \nonumber \\
   &= (1 - \gamma_k)^2 (1+\gamma_k)^2 \EX[\|\frac{1}{1+\gamma_k}\big (\vect{v}_{k-1} - \vect{g}(\vect{\theta}_{k-1}) \big) + \frac{\gamma_k}{1+\gamma_k}\eta_k \vect{p}_k\|^2] + \gamma_k^2\EX[ \| (\hat{\vect{g}}(\vect{\theta}_k) - \vect{g}(\vect{\theta}_k))\|^2].
\end{align}
Using the result from Lemma \ref{lem::lemm-variance-FG} in \eqref{eq::13}, we have 
\begin{align}\label{eq:::16}
        \EX&[\|\vect{v}_k - \vect{g}(\vect{\theta}_k)\|^2] \leq (1 - \gamma_k)^2 (1+\gamma_k)^2 \EX[\|\frac{1}{1+\gamma_k}\big (\vect{v}_{k-1} - \vect{g}(\vect{\theta}_{k-1}) \big) + \frac{\gamma_k}{1+\gamma_k}\eta_k \vect{p}_k\|^2] + \gamma_k^2 \mathsf{G}.
\end{align}
Subsequently, invoking Jensen's inequality in \eqref{eq:::16}, we have 
\begin{align}
    \EX&[\|\vect{v}_k\! -\! \vect{g}(\vect{\theta}_k)\|^2] \!\leq\! (1 \!-\! \gamma_k)^2 (1+\gamma_k) \EX[\|\vect{v}_{k-1} \!- \vect{g}(\vect{\theta}_{k-1}) \|^2] + (1 - \gamma_k)^2 (1+\gamma_k) \gamma_k \eta_k^2  + \gamma_k^2 \mathsf{G} \nonumber \\
   & { \stackrel{\text{(a)}}{\leq} } (1 - \gamma_k) \EX[\|\vect{v}_{k-1} - \vect{g}(\vect{\theta}_{k-1}) \|^2] + (1 - \gamma_k) \gamma_k \eta_k^2  + \gamma_k^2 \mathsf{G} \nonumber\\
   & { \stackrel{\text{(b)}}{\leq} }(1 - \gamma_k) \EX[\|\vect{v}_{k-1} - \vect{g}(\vect{\theta}_{k-1}) \|^2] + \gamma_k \eta_k^2  + \gamma_k^2 \mathsf{G} \label{eq::45},
\end{align}
RHS of inequality (a) in \eqref{eq::45} follows from $(1 - \gamma_k)^2 (1+\gamma_k) = (1-\gamma_k)(1-\gamma_k^2) \leq (1-\gamma_k)$, RHS of inequlity (b) in \eqref{eq::45} follows from $(1-\gamma_k) \leq 1$. Denoting $\delta_k = \EX[\|\vect{v}_k - \vect{g}(\vect{\theta}_k)\|^2]$, we have $ \delta_k \leq (1 - \gamma_k) \delta_{k-1} + \gamma_k\eta_k^2  + \gamma_k^2\mathsf{G}.$

Invoking Comparison Lemma~\cite{HKK:02} and Lyapunov convergence analysis, we consider $V_{k-1} = |d_{k-1}|$ as our candidate Lyapunov~function to write
\begin{align*}
    V_{k} - V_{k-1} &= |(1 - \gamma_k) d_{k-1} + \gamma_k\eta_k^2  + \gamma_k^2\mathsf{G}| - |d_{k-1}|\nonumber \\ 
   & \leq - \gamma_k |d_{k-1}| + \gamma_k\eta_k^2 + \gamma_k^2\mathsf{G} \nonumber \\
   &= -\beta \gamma_k |d_{k-1}| - (1-\beta)\gamma_k |d_{k-1}| + \gamma_k\eta_k^2 + \gamma_k^2\mathsf{G}
\end{align*}
where $0< \beta < 1$. Then if $|d_{k-1}| \geq \frac{\eta_k^2 + \gamma_k\mathsf{G}}{(1-\beta)}$, we have exponentially convergence to the ball with radius $\frac{\eta_k^2 + \gamma_k\mathsf{G}}{(1-\beta)}$. Then, the overall rate of the convergence of  $\lim_{k\to\infty} \EX[\|\vect{v}_k - \vect{g}(\vect{\theta}_k)\|^2] = 0$ is determined by \mbox {$\min\{$$O(\eta_k^2)$, $O(\gamma_k)$\}}. 
\end{proof}

\begin{proof}[Proof of Theorem~\ref{thm::thm2}]
Following the same procedure as in the proof of Theorem~\ref{thm::thm1}, we obtain $ \mathbb{E} [f(\mathbf{x}_{k+1}) - f(\vect{\theta}^{\star})] \leq (1 - \alpha_k) \mathbb{E} [f(\mathbf{x}_k)\! -\! f(\vect{\theta}^{\star})] 
     \!+\! \alpha_k D \sqrt{\mathbb{E} [\|\vect{g}(\vect{\theta}_{k}) - \mathbf{v}_k\|^2]} + \frac{M \alpha_k^2 D^2}{2}.$
By defining $\delta_{k+1} = \EX [f(\vect{\theta}_{k+1}) - f(\vect{\theta}^{\star})]$ we can~write $ \delta_{k+1} \leq (1-\alpha_k)\delta_k +  \psi_1\alpha_k +\psi_2 \alpha_k^2 ,~  \forall k \geq 0 $
where  $\psi_1 =D \sqrt{\EX [\|\vect{g}(\vect{\theta}_k) - \vect{v}_k\|]^2} $ and $\psi_2 = \frac{M D^2}{2}$. Same as the proof in theorem \ref{thm::thm1}, choosing $V_k = |\delta_k|$ as our candidate Lyapunov function, we obtain
\begin{align*}
    V_{k+1} - V_k &\leq -\beta \alpha_k |\delta_{k}| - (1-\beta) \alpha_k |\delta_{k}| + \alpha_k(\psi_1 + \psi_2\alpha_k),
\end{align*}
where $0< \beta < 1$. Then if $|\delta_{k}| \geq \frac{\psi_1 + \psi_2\alpha_k}{1-\beta}$, we have exponentially convergence to the ball with radius $\frac{\psi_1 + \psi_2\alpha_k}{1-\beta}$. Hence, using Lemma \ref{lem:averaged}, the overall rate of convergence of $\lim_{k\to\infty} \EX [f(\vect{\theta}_{k}) - f(\vect{\theta}^{\star})] = 0$ is determined by 
{$\min \{O(\frac{\alpha_k}{\gamma_k}), O(\gamma_k)\}$}.
\end{proof}

\begin{proof}[Proof of Theorem~\ref{thm::thm2_notcon}]
Under Assumption \ref{assump::smoothness}, we can write
$\EX[f(\vect{\theta}_{k+1})|\vect{\theta}_{k}] \leq f(\vect{\theta}_{k})+\big \langle\vect{g}(\vect{\theta}_{k}), \EX[\vect{\theta}_{k+1} - \vect{\theta}_{k}|\vect{\theta}_{k}] \big \rangle +\frac{M}{2}\EX\big[\|\vect{\theta}_{k+1} - \vect{\theta}_{k}\|^2\big| \vect{\theta}_{k}\big]$. Substituting the update rule (line 8) of Algorithm \ref{alg::AFG-FW} in this relationship leads to
\begin{align}
&\!f\left(\vect{\theta}_{k+1}\right) \! \leq \!f\left(\vect{\theta}_k\right)\!+\!\left\langle\vect{g}(\vect{\theta}_{k}), \vect{\theta}_{k+1}-\vect{\theta}_k\right\rangle\!+\!\frac{M}{2}\left\|\vect{\theta}_{k+1}-\vect{\theta}_k\right\|^2 \nonumber \\
& = f\left(\vect{\theta}_k\right)+\left\langle\vect{g}(\vect{\theta}_{k}), \alpha_k\left(\vect{s}_k-\vect{\theta}_k\right)\right\rangle +\frac{M}{2}\left\|\alpha_k\left(\vect{s}_k-\vect{\theta}_k\right)\right\|^2 \nonumber\\ \label{eqq1}
& \stackrel{\text{(a)}}{\leq}  f\left(\vect{\theta}_k\right)+\alpha_k\left\langle\vect{g}(\vect{\theta}_{k}), \vect{s}_k-\vect{\theta}_k\right\rangle\!+\!\frac{M D^2 \alpha_k^2}{2},
\end{align} 
RHS of inequality (a) in \eqref{eqq1} follows from Assumption \ref{asm:comapct}. Adding and subtracting $\alpha_k\left\langle\vect{v}_k, \vect{s}_k-\vect{\theta}_k\right\rangle$ into \eqref{eqq1} results in
\begin{align}\label{eq::46}
& f\left(\vect{\theta}_{k+1}\right) \leq f\left(\vect{\theta}_k\right)+\alpha_k\left\langle\vect{v}_k, \vect{s}_k-\vect{\theta}_k\right\rangle   +\alpha_k\left\langle\vect{g}(\vect{\theta}_{k})-\vect{v}_k, \vect{s}_k-\vect{\theta}_k\right\rangle +\frac{M D^2 \alpha_k^2}{2} \nonumber \\
&  { \stackrel{\text{(a)}}{\leq}} f\left(\vect{\theta}_k\right)-\alpha_k \mathcal{K}\left(\vect{\theta}_k\right)+D \alpha_k\left\|\vect{g}(\vect{\theta}_{k})-\vect{v}_k\right\|+\frac{M D^2 \alpha_k^2}{2}.\nonumber
\end{align}
RHS of inequlity (a) in \eqref{eq::46} follows from Assumption \ref{asm:comapct}, use of the definition of $\mathcal{K}\left(\vect{\theta}_k\right)$ and Cauchy-Schwarz inequality. Taking total expectations on both sides, we obtain
\begin{align}
&\EX\left[f\left(\vect{\theta}_{k+1}\right)\right] \leq \EX\left[f\left(\vect{\theta}_k\right)\right]-\alpha_k \EX\left[\mathcal{K}\left(\vect{\theta}_k\right)\right] +D \alpha_k \EX [\left\|\vect{g}(\vect{\theta}_{k})-\vect{v}_k\right\|] 
+\frac{M D^2 \alpha_k^2}{2} \\
&\leq \EX\left[f\left(\vect{\theta}_k\right)\right]-\alpha_k \EX\left[\mathcal{K}\left(\vect{\theta}_k\right)\right]+D \alpha_k \sqrt{\EX [\left\|\vect{g}(\vect{\theta}_{k})-\vect{v}_k\right\|^2]} + \frac{M D^2 \alpha_k^2}{2} 
\nonumber \\ &\leq \EX\left[f\left(\vect{\theta}_k\right)\right]-\alpha_k \EX\left[\mathcal{K}\left(\vect{\theta}_k\right)\right]
+O(\frac{1}{k^{1.25}}) + O(\frac{1}{k^{2}}),\nonumber 
\end{align}
The last summands are used as a shorthand notation, for brevity, to represent decaying terms (Recall Lemma~\ref{lem::variance_convergence}). 
Rearranging and summing $k$ from $1$ to $K$, we~have 
\begin{align}\label{eqq:2}
&\frac{\sum_{k=1}^{K} \alpha_k \EX \left[\mathcal{K}\left(\vect{\theta}_k\right)\right]}{\sum_{k=1}^{K} \alpha_k}  \leq \frac{f\left(\vect{\theta}_0\right)-\EX\left[f\left(\vect{\theta}_{K + 1}\right)\right]}{\sum_{k=1}^{K} \alpha_k} + \frac{\sum_{k=1}^{K} O(\frac{1}{k^{1.25}})}{\sum_{k=1}^{K} \alpha_k} 
+ \frac{\sum_{k=1}^{K} O(\frac{1}{k^{2}})}{\sum_{k=1}^{K} \alpha_k}  \\
& \quad\leq \frac{f\left(\vect{\theta}_0\right)-f\left(\vect{\theta}^\star\right)}{\sum_{k=1}^{K} \alpha_k}  + \frac{\sum_{k=1}^{K} O(\frac{1}{k^{1.25}})}{\sum_{k=1}^{K} \alpha_k} + \frac{\sum_{k=1}^{K} O(\frac{1}{k^{2}})}{\sum_{k=1}^{K} \alpha_k},\nonumber 
\end{align}
where both $\sum_{k=1}^{K} O(\frac{1}{k^{1.25}})$ and $\sum_{k=1}^{K} O(\frac{1}{k^{2}})$ are convergent series which can be verified using the integral test. From \eqref{eqq:2} and substituting $\sum_{k=1}^{K} \alpha_k \approx \log K$ for large $K$, we have 
\begin{align}
\EX \left[\mathcal{K}\left(\vect{\theta}_K\right)\right] & \leq \frac{f\left(\vect{\theta}_0\right)-f\left(\vect{\theta}^\star\right)}{\log K}  + E
\end{align}
where $E$ is of order $O(\frac{1}{\log K})$, which concludes the proof.
\end{proof}


\end{document}